%% file: main.tex
\documentclass{article}

\usepackage[final]{corl_2018} 
\usepackage{multicol}
\usepackage{float}
\usepackage{mathtools}
\usepackage{amsfonts}
\usepackage{booktabs}
\usepackage{epstopdf}
\usepackage{svg}
\usepackage[numbers]{natbib}
\usepackage{multirow}
\usepackage{grffile}
\usepackage[noend]{algpseudocode}
\usepackage{algorithm}
\usepackage{amsmath}
\usepackage{enumitem}
\usepackage{amsfonts}
\usepackage[capitalise]{cleveref}

\newcommand{\DeterAgg}{DRAG }
\newcommand{\DeterAggns}{DRAG}

\title{Learning Neural Parsers with \\Deterministic Differentiable Imitation Learning} 

\author{
  Tanmay Shankar\\
  Robotics Institute \\
  Carnegie Mellon University\\
  Pittsburgh, PA 15213 \\
  \texttt{tanmayshankar@cmu.edu} \\
    \And  
  Nicholas Rhinehart\\
  Robotics Institute\\
  Carnegie Mellon University\\
  Pittsburgh, PA 15213 \\
  \texttt{nrhineha@cs.cmu.edu} \\
    \AND
Katharina Muelling\\
  Robotics Institute\\
  Carnegie Mellon University\\
  Pittsburgh, PA 15213 \\
  \texttt{kmuelling@nrec.ri.cmu.edu} \\
    \And
     Kris M. Kitani  \\
  Robotics Institute\\
  Carnegie Mellon University\\
  Pittsburgh, PA 15213 \\
  \texttt{kkitani@cs.cmu.edu} \\
  }

\begin{document}
\maketitle
    

\begin{abstract}

We explore the problem of learning to decompose spatial tasks into segments, as exemplified by the problem of a painting robot covering a large object. Inspired by the ability of classical decision tree algorithms to construct structured partitions of their input spaces, we formulate the problem of decomposing objects into segments as a parsing approach. We make the insight that the derivation of a parse-tree that decomposes the object into segments closely resembles a decision tree constructed by ID3, which can be done when the ground-truth available. 
We learn to imitate an expert parsing oracle, such that our neural parser can generalize to parse natural images without ground truth. We introduce a novel deterministic policy gradient update, DRAG (i.e., DeteRministically AGgrevate) in the form of a deterministic actor-critic variant of AggreVaTeD \cite{Sun2017}, to train our neural parser. From another perspective, our approach is a variant of the Deterministic Policy Gradient \cite{Silver2014, Lillicrap2015} suitable for the imitation learning setting. The deterministic policy representation offered by training our neural parser with DRAG allows it to outperform state of the art imitation and reinforcement learning approaches. 
\end{abstract}



\input{0_intro2.tex}

\input{1_relatedwork.tex}
\input{2_method.tex}

\input{3_newresults.tex}
\input{4_conclusion.tex}

\clearpage

\acknowledgments{The authors would like to thank Wen Sun, Anirudh Vemula, and Arjun Sharma for technical discussions, and Marinus Analytics for providing us access to computing resources for our experiments. We would also like to thank the anonymous reviewers for their comments on our paper.}

\bibliographystyle{plainnat}
\bibliography{references}  

\clearpage
\input{supplementary_body.tex}

\end{document}

%% file: 0_intro2.tex
\vspace*{-1em}
\section{Introduction}
\vspace*{-1em}
Consider the task of a robot painting an object or an aerial robot surveying a large field. These spatial tasks represent a coverage problem that the robot may not be able to address in a single shot. For instance, a robot may not be able to paint the entirety of a large object with a single stroke, being limited by the footprint of its paint brush. Instead, the robot must decompose the spatial task of painting objects into smaller segments that it can cover in single stroke.
However, discovering or learning an appropriate decomposition of such tasks into segments is challenging. In the object-painting problem, there may be several constraints upon the resultant segments, such as the overall paint coverage. 
Further, there exist multiple ways to decompose an object into constituent segments - for example, an object may be decomposed into length or breadth-wise segments. 

A few well studied algorithms are able to somewhat circumvent these challenges. In particular, 
when applied to the task of classifying a set of points, ID3 \cite{Quinlan86inductionof} and C4.5 \cite{Salzberg1994} recursively partition the input space in order to achieve an accurate classification of the points.
As we demonstrate in \cref{sec:method}, an ID3-like Information Gain maximizing algorithm, IGM, builds a similar partitioning of an object as in \cref{fig:teaser}. However, this IGM algorithm (like the ID3 and C4.5 algorithms it is derived from) requires ground truth classification labels, and 
cannot be applied to decompose novel objects for which the ground truth labels are unavailable, or are expensive to obtain.

To bypass this issue, we approach the problem of decomposing objects into segments as a \textit{parsing} problem, based on the insight that the derivation of a parse-tree (\cref{fig:teaser}) that decomposes a given object into segments closely resembles a decision tree constructed by IGM (\cref{fig:teaser}). 
Rather than learn to parse objects by reinforcement learning (RL) as in \cite{conf/cvpr/TeboulKSKP11}, we propose to learn how to parse objects into such structured decompositions via \textit{imitation learning} (IL), treating the IGM algorithm as a \textit{parsing oracle}. 
Our neural parser is trained to parse objects by imitating the IGM oracle \textit{observing only raw object images} as input, while the IGM oracle exploits access to ground truth information to demonstrate how to parse a particular object. This allows our neural parser to construct structured decompositions of novel unseen objects despite lacking ground truth information. As expected, our imitation learning approach significantly outperforms reinforcement learning baselines in practice.


We further introduce a novel deterministic policy gradient update, \DeterAggns, suitable for training deterministic policies in the hybrid imitation-reinforcement learning setting. 
The \DeterAgg policy gradient update serves to train the deterministic policy component of our neural parser, eliminating the complexity of maintaining probability distributions specific to the parsing setting. 
By rephrasing the AggreVaTeD \cite{Sun2017} objective in the deterministic policy case, we retrieve a gradient update to a deterministic policy that relies on a differentiable approximation to the oracle's cost-to-go. 
This policy gradient update may be viewed as a \textit{deterministic actor-critic variant of AggreVaTeD}, 
which we refer to as \DeterAgg (i.e., DeteRministically AGgrevate). \DeterAgg may also be viewed as a variant of the Deterministic Policy Gradient \cite{Lillicrap2015} suitable for imitation learning, and replaces an approximation to the true gradient in the original Deterministic Policy Gradient \cite{Lillicrap2015}, with the true policy gradient. 

Training our parser via \DeterAgg allows our parser to outperform several baselines on the task of parsing novel objects, showcasing its potential to achieve performance closer to that of the oracle than several other existing imitation and reinforcement learning approaches.

%% file: 1_relatedwork.tex
\vspace*{-1em}
\section{Related Work}
\label{relatedwork}
\vspace*{-1.em}
\textbf{Facade Parsing:} Facade parsing attempts to identify the topology of a building facade, by parsing an image of the facade into its various components \cite{conf/cvpr/TeboulSKP10, conf/cvpr/TeboulKSKP11,martinovic2012three,zhang2013layered,kozinski2015mrf}.
\cite{conf/cvpr/TeboulSKP10, conf/cvpr/TeboulKSKP11} learn to apply production rules of a grammar to reduce a shape into its constituent segments in the RL setting. We build on \cite{conf/cvpr/TeboulKSKP11}, addressing the problem of decomposing objects. In contrast to other facade parsing approaches that use labels of the \emph{resulting parse} \cite{martinovic2012three,zhang2013layered,kozinski2015mrf}, we seek to imitate the \textit{decisions} of an expert parser.


\textbf{Policy Gradient Reinforcement Learning:}
Stochastic policy gradient approaches \cite{sutton2000policy, Williams1992, DBLP:journals/corr/abs-1205-4839, jie2010connection} have been used to learn control policies in the reinforcement learning (RL) setting.
\citet{Silver2014} introduced the Deterministic Policy Gradient (DPG), a deterministic counterpart to \cite{sutton2000policy}, and later extended DPG to the function approximator case \cite{Lillicrap2015}. We introduce a variant of DPG \cite{Silver2014, Lillicrap2015} suitable for imitation learning, that removes the approximation of the true gradient used in DPG \cite{Silver2014, Lillicrap2015}. 

\textbf{Imitation Learning:}
Recent imitation learning algorithms \cite{daume2009search, ross2011reduction, Ross2011} address the setting when one has access to an expert policy that may be queried. \citet{Ross2011} demonstrated an \textit{interactive} imitation learning paradigm, DAgger, is preferable over a naive behavioural cloning approach.
\citet{Ross2014} further introduced AggreVaTe, using estimates of the cost-to-go of the expert to better learn control policies. \citet{Sun2017} subsequently derived a stochastic policy gradient update of the AggreVaTe \cite{Ross2014} objective, enabling its use on complex neural network policies. 
\cite{sun2018truncated, 2018arXiv180510413C} further explore the hybrid imitation and reinforcement learning, by reward shaping using an oracle and switching to policy gradient RL after imitation respectively.
We follow \cite{Ross2014, Sun2017}, by training agents with partial information to imitate oracles with full information at train time, as in \cite{Choudhury2017, bhardwaj2017learning}. 

\textbf{Semantic Segmentation:} 
The problem of semantic segmentation addresses assigning semantic labels to pixels in a given image. 
\cite{conf/cvpr/LiWC12} employed a graph partitioning approach for image segmentation, to construct image superpixels. 
Several recent works \cite{DBLP:journals/corr/LongSD14, DBLP:journals/corr/BadrinarayananH15, conf/cvpr/LiWC12, sharma2015deep}
train end-to-end models on large scale datasets for semantic segmentation, as more thoroughly reviewed in \cite{DBLP:journals/corr/Garcia-GarciaOO17}. 
While the notion of a set of segments with assigned labels is common to our parsing approach and semantic segmentation, our approach builds a hierarchical decomposition of the object image as the end result. Our problem further differs in that constraints may be imposed that exclude arbitrary results such as the aforementioned stroke coverage constraints of a painting robot.

%% file: 2_method.tex
\begin{figure*}[tb]
	\centering
	\includegraphics[width=0.7\linewidth]{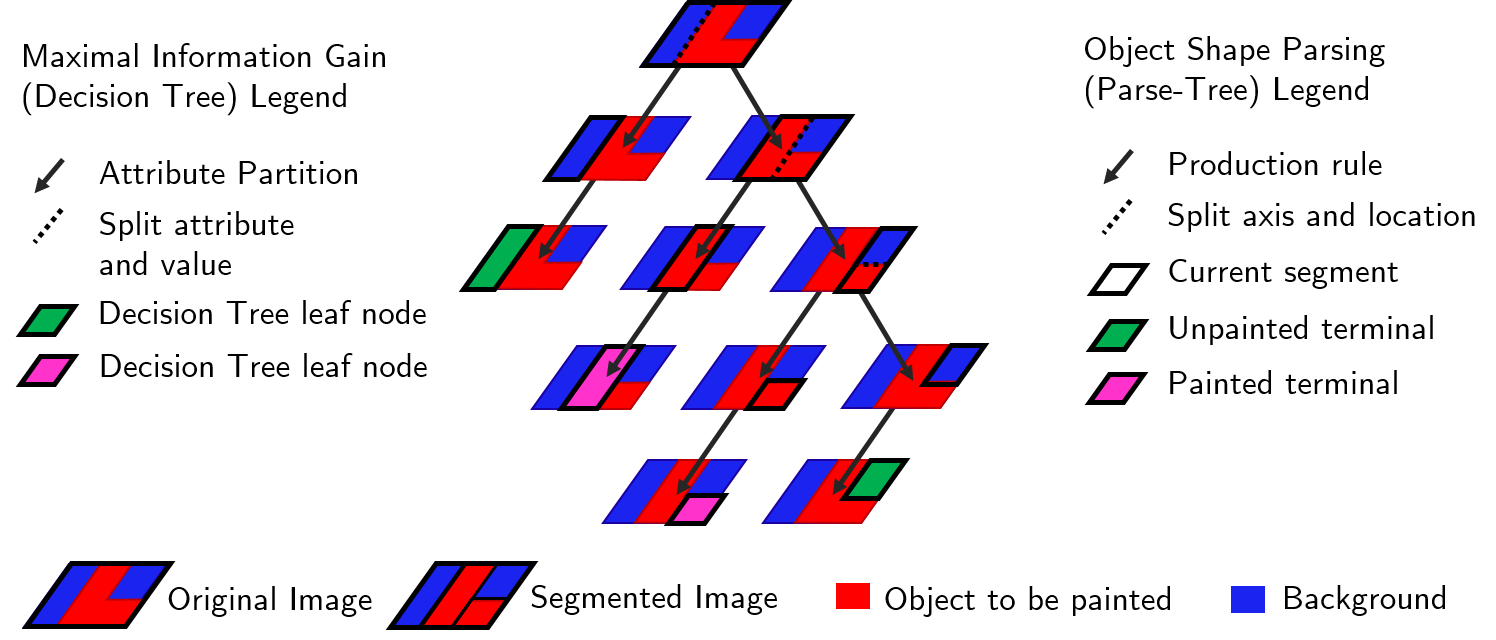}
	\vspace*{-0.5em}
	\caption{ \small Constructing an equivalent hierarchical decomposition of an object image by two methods. The Information Gain Maximization algorithm creates a decision tree (legend on the left), while the Shape Parsing approach to the right constructs a parse tree (legend to the right).
	The equivalent hierarchical decomposition (center), shows correspondence of each node in the decision tree and parse tree to an image segment.}
	\label{fig:teaser}
	\vspace*{-1.5em}
\end{figure*}
\vspace*{-1em}
\section{Method}
\vspace*{-1em}
\label{sec:method}
We seek to learn how to parse objects directly using object images as input, by learning to imitate an expert parsing oracle.
The connection between object parsing and the decision trees constructed by IGM \cref{fig:teaser} afford us such a parsing oracle, that makes the imitation setting preferable over the RL setting followed in \cite{conf/cvpr/TeboulKSKP11}. 
To successfully learn a policy capable of imitating a parsing oracle, we introduce \DeterAggns, a deterministic actor-critic variant of AggreVaTeD \cite{Sun2017}.
We explain our approach to learning this parsing policy by first describing the problem setting of parsing objects, then describing the IGM algorithm that serves as a ground truth parsing oracle. We then demonstrate how the object parsing problem may be framed as an MDP, finally highlighting how we use \DeterAgg to train a neural parser to decompose objects. 

Consider the following problem setting. Given an object image to be painted, our objective is to decompose the object image into a set of segments, maximizing the paint coverage of the object while minimizing the ``paint wasted''. 
We may view this as assigning labels of whether or not to paint each pixel, where painting object pixels increases coverage, while not painting the surroundings decreases ``wasted paint''. Under this label assignment problem, our objective translates to constructing an object decomposition (a set of segments) constrained to mutually compose the object image, while \textit{accurately} assigning labels of whether or not to paint each of the resultant segments. 

\vspace*{-1em}
\subsection{Parsing Objects by Imitating Maximal Information Gain}
\vspace*{-1em}
Representing the painting process as a labeling problem allows us to employ decision tree algorithms such as ID3 \cite{Quinlan86inductionof} and C4.5 \cite{Salzberg1994}, that naturally address labeling tasks by partitioning the space (i.e., the object image) into regions that each contain pixels of a single class. 
ID3 achieves this by using ground truth information to select the partitioning with the maximum information gain over the resultant segments. By modifying ID3 to allow multiple splits (i.e., partitions) along a particular axis (or attributes in ID3), we construct a ground-truth oracle that is able to perfectly label any object image allowed sufficient partitions. We refer to this oracle as the Information Gain Maximization algorithm (IGM), or $\pi_{\rm IGM}$.
Fig. \ref{fig:teaser} depicts such a ``decision tree'' constructed by IGM on a toy image. 

The IGM oracle uses ground truth labeling of object images, which are unavailable for novel objects. 
To address this issue, we draw inspiration from facade parsing literature \cite{conf/cvpr/TeboulSKP10, conf/cvpr/TeboulKSKP11}, where images of facades were decomposed into various components via \textit{shape-parsing}. We observe that parse-trees that decompose an image into segments in \cite{conf/cvpr/TeboulKSKP11} resemble the decision trees constructed by IGM (Fig. \ref{fig:teaser}). Motivated by this insight, we decompose objects into segments via a \textit{parsing} approach. 



\vspace*{-1em}
\subsection{Shape Parsing Objects}
\vspace*{-1em}
To learn how to decompose object images into their constituent segments, we require an appropriate representation of the recursive object decompositions that arise in the divide-and-conquer paradigm. 
Shape parsing \cite{conf/cvpr/TeboulKSKP11} provides us a compact representation of such potentially complex decompositions of an object. 
We adopt a \textit{shape parsing} approach similar to \cite{conf/cvpr/TeboulKSKP11}, using a binary split grammar to represent the hierarchical object decomposition. Formally, we use a probabilistic context-free grammar $\mathcal{G}$, defined as a tuple $\mathcal{G} = (\mathcal{V}, \mathcal{T}, \mathcal{R}, V_0, \mathcal{P})$, where $\mathcal{V}$ is a set of non-terminal symbols, $\mathcal{T}$ is a set of terminal symbols, $V_0$ is a starting symbol, $\mathcal{R}$ is a set of production rules, and $\mathcal{P}$ defines a set of probabilities of applying the production rules on a given non-terminal $V$.

\noindent\textbf{Set of non-terminal symbols} $\mathcal{V}$:
A non-terminal symbol $V \in \mathcal{V}$ is defined as an axis-aligned rectangle over the input image. Each non-terminal is specified with a set of attributes $(x,y,w,h)$, where $(x,y)$ defines the origin of the rectangle and $(w,h)$ define the spatial extents of the rectangle along the horizontal and vertical directions respectively. The starting symbol $V_0 \in \mathcal{V}$ is a rectangular region encompassing the entire object image.

\noindent\textbf{Set of terminal symbols} $\mathcal{T}$:  A terminal symbol $T \in \mathcal{T}$ is an axis-aligned rectangle image segment, with an additional attribute $b$ denoting whether a region is to be painted or not. 

\noindent\textbf{Set of production rules} $\mathcal{R}$: We consider binary split rules, which split a non-terminal $V$ along its axes into two constituent non-terminals. A split rule is specified by a horizontal or vertical split axis ($h$ or $v$), and split location $l$, as $V  \xrightarrow[]{h:l} V~V$, or $V \xrightarrow[]{v:l} V~V$. An instance of a horizontal split rule is visualized in the top row of Fig. \ref{fig:teaser}, resulting in two non-terminals ($V_{\textrm{left}}$, $V_{\textrm{right}}$). 
Our grammar also includes assignment rules that assign a precedant non-terminal symbol to one of the two terminal symbols, a region $b_{\rm{p}}$ to be painted, or a region not to be painted, $b_{\rm{np}}$. The full set $\mathcal{R}$ is as follows:
\begin{align}
\mathcal{R} &= \{ 
V_0 \rightarrow V \nonumber,  
V \xrightarrow{x:l} (V_{\textrm{left}}, V_{\textrm{right}}) \nonumber,
V \xrightarrow{y:l} (V_{\textrm{top}}, V_{\textrm{bottom}}) \nonumber,
V \rightarrow b_{\rm{p}} \nonumber,
V \rightarrow b_{\rm{np}} \nonumber \}
\end{align}
\noindent\textbf{Rule probabilities} $\mathcal{P}$:
Production rules $r$ have an associated probability $p_r$ of applying rule $r$ on the current non-terminal (i.e., an image segment). In our problem, we seek to learn these probabilities $p_r$ of applying the production rules, along with the associated split location attribute $l$ of the rules.

Recursively applying production rules $\mathcal{R}$ of the grammar $\mathcal{G}$ described above on an object image $I$ (and the resultant segments) decomposes the image into its constituent segments, a process known as shape-parsing. 
Parsing an object image results in a hierarchical decomposition of the object, or an object parse-tree, as depicted for a toy object image in Fig. \ref{fig:teaser}.  

Starting with the entire image (represented by starting symbol $V_0$), the object parse-tree is constructed by expanding each node in the tree in a top-down and depth-first manner. While expanding a node $N$ of the parse-tree, we sample a rule $r \in \mathcal{R}$ with probability $p_r$, from the set of rules applicable on $N$. We add the antecedants of this rule to the parse tree as children nodes of the expanded node. We then continue to derive the tree in a depth-first fashion, moving on to the next unexpanded node in the tree. The labels assigned to these leaf nodes of a fully expanded image parse tree yield a segmentation of the image. We present a sample object image, the corresponding image parse tree derived via shape parsing, and the final segmented image in Fig. \ref{fig:teaser}. 

\vspace*{-1em}
\subsection{Shape Parsing as a Markov Decision Process}
\label{framingasrl}
\vspace*{-1em}
The shape parsing process may be seen as a sequential decision making process, where a parsing agent finds a sequence of partitions and label assignments that maximizes the paint coverage of the object, while affording us a decomposition of the object. 
We can formally describe the sequential process of shape parsing as a Markov Decision Process (MDP) $\mathcal{M}$. 

Here, the current image segment $\rho_t$ corresponds to node $N$ in the parse tree as the current state $s \in S$ of $\mathcal{M}$. Actions $a \in \mathcal{A}$ correspond to applying production rules $r \in \mathcal{R}$ with a particular split location $l \in [0, 1]$. Upon taking an action $a$ from state $s$, we ``transition'' to the next unexpanded node $s'$ in the parse tree. Here, $s'$ is specified by the deterministic transition dynamics $p(s_{t+1}|a_t,s_t)$ enforced by the top-down, depth-first expansion of the tree $\tau$. 

The sequence of nodes expanded during the expansion of the image parse-tree corresponds to the sequence of states observed by our agent. This may be incorporated elegantly in the definition of both the one-step reward and the cumulative discounted reward (returns) $G$ of the agent. The one-step reward function encodes the coverage objective of our object parsing problem.
For every terminal symbol  
$T$, we evaluate the image correlation between the predicted label assignments $\mathbb{P}$ over each of the terminal segments, and the ground truth paint labels of the objects $\mathbb{L}$. 
For any non-terminal node $V$ in the parse tree, the return $G(V)$ is defined \textit{recursively} in \cref{eq:returns} as the discounted sum of the returns of all child nodes of $V$. This recursion propagates rewards up the tree in a bottom-up manner, starting from the terminal leaf-nodes $T$, where the return is the one-step reward. 
\begin{equation}
    G( N ) = \begin{cases} \sum_{(x,y) \in N} \mathbb{L}(x,y) \mathbb{P}(x,y) & \text{if}~N \in \mathcal{T} \\
    \sum_{c \in \textrm{Children}(N)} G(c) & \text{if}~N \in \mathcal{V}
    \end{cases}
    \label{eq:returns}
\end{equation}
Here, $x$ and $y$ represent pixel locations in the image, $\rm{Children}(N)$ is the set of children nodes of $N$ in the tree, $C$ indexes these child nodes.



We seek to learn a policy $\pi : s \rightarrow a $ mapping the current state $s$ to one of the possible actions $a$ available in the current state. 
In our setting, the policy must select (1) which production rule to apply, and (2) a corresponding split location. 
To predict these facets of an action $a$ directly from visual input of the current image segment, we represent our policy as a deep convolutional network. 
Our policy network thus takes in as input an image region $\rho_t(x,y,w,h)$ defined by the current non-terminal $V$ that is being expanded.

The policy then predicts (1) a categorical probability distribution over the valid production rules, $\pi(r|s_t, \theta)$, and (2) a split-location $l = \mu(s_t|\theta)$ within the current image segment in case of applying a split rule. 
Since the size of the image segment varies at every step, the valid split locations \textit{vary} with the current state. While maintaining a valid probability distribution over such a varying range of values is possible, it is notably challenging, due to the normalizing a distribution across changing scale and limits of the distribution at every step. Instead, we employ a \textit{deterministic} representation of the split location policy, thus $\mu(s_t|\theta)$ is \textit{deterministically} predicted as a scaled logistic function of the deep network features. 
The two components of our policy network represent the \textit{mixed} deterministic and stochastic nature of the our policy. 

\vspace*{-0.5em}
\subsection{Learning the Shape Parser via Imitation Learning}
\vspace*{-1em}
While \citet{conf/cvpr/TeboulKSKP11} learn a shape-parser via reinforcement learning, it is known that the imitation learning paradigm is preferable to reinforcement learning if an expert agent may be easily obtained \cite{Ross2014}. In our case, the connection object parsing and the decision trees constructed by IGM afford us such an expert. We hence consider 
learning this mixed deterministic-stochastic policy in the imitation learning setting. The stochastic component of the policy $\pi(r|s, \theta)$ may be learned via \textit{off-policy} Monte-Carlo \cite{Williams1992} or actor-critic \cite{DBLP:journals/corr/abs-1205-4839} policy gradient algorithms (we point the reader towards \cite{Silver2014} for a review of these algorithms). However, existing algorithms for learning the deterministic component of the policy $l = \mu(s|\theta)$ (notably the Deterministic Policy Gradient introduced in \cite{Silver2014}), have only been developed in the reinforcement learning setting, not the imitation learning setting. 

To learn the deterministic component of the policy $\mu(s|\theta)$, we introduce a deterministic policy gradient update suitable for training deterministic policies in the cost sensitive imitation learning setting. \DeterAgg (DeteRministically AGgrevate) may be viewed as a deterministic actor-critic variant of AggreVaTeD \cite{Sun2017}, or alternatively, a variant of the Deterministic Policy Gradient \cite{Lillicrap2015} suitable for imitation learning. \DeterAgg replaces an approximation to the true gradient in the original Deterministic Policy Gradient \cite{Lillicrap2015}, with the correct gradient. 


We present \DeterAgg by first describing the AggreVaTe / AggreVaTeD setting - an ideal starting point given we have an oracle (IGM) that we may \textit{query} for the optimal action to execute from any state. 
AggreVaTe \cite{Ross2014} and AggreVaTeD \cite{Sun2017} approach the problem of learning a policy $\pi_{\theta}$ by training the policy $\pi_{\theta_{n}}$ at training iteration $n$ to minimize the cost-to-go $Q^*$ of the oracle $\pi^*$, over the aggregated distribution of states $d_{\pi_n}^t$ induced by the current learner's policy, $\pi_{\theta_{n}}$. To do so, they roll-out a trajectory with a mixture policy $\pi_n(s) = \beta \pi^*(s)  + (1-\beta) \pi_{\theta}(s)$ till time step $t \in [1,...,H]$, and subsequently follow the expert $\pi^*$ then onwards. $\beta$ simply represents the mixing coefficient, and $H$ is the horizon length of the MDP, and the aggregated distribution of states $d_{\pi_n}^t$ is defined as $\sum_{ \{ s_i, a_i\}_i \leq t-1 } \rho_0 (s_1) \Pi_{i=1}^{t-1} \pi_n(a_t|s_{t-1}) p(s_t |s_{t-1},a_{t-1}) $, and $\rho_0$ is the initial state distribution. 
The AggreVaTe \cite{Ross2014} objective to be optimized may be represented as:
\begin{equation}
J_n(\theta) = \mathbb{E}_{\ t \sim U(1,...,H), s_t \sim d_{\pi_{1:n}}^t, a_t \sim \pi_n(a|s)} \Big[ Q_t^*(s_t,a_t) \Big].
\end{equation}
\citet{Sun2017} assume a stochastic policy $\pi(a|s,\theta)$ to derive a stochastic policy gradient update to the parameters of the policy $\theta$.
\setlength{\textfloatsep}{0.1cm}
\renewcommand{\algorithmicrequire}{\textbf{Input:}}
\renewcommand{\algorithmicensure}{\textbf{Output:}}
\begin{algorithm}[t]
	\caption{Train Parser via \DeterAgg}
	\label{alg}
	\begin{algorithmic}[1]	
	    \Require $\mathcal D, \pi^*, \beta, N_\text{iterations}$ \Comment{Require a dataset, expert parser, mixing parameter, iterations}
	    \Ensure $\pi_{\theta}$ \Comment{Output the learned policy}
		
		\State ${\theta} \gets \mathbf{0}, \mathcal M \gets \{\}$ \Comment{Initialize Policy Parameters,  Initialize Memory}
		\For {$i \in [1,2,...,N_{\rm iterations}]$}
		\State $\pi_i \gets \beta \pi^* + (1-\beta) \pi_{\theta}$
		\For {$j \in [1,2,..., N_{\rm images} = |\mathcal D|]$}
		\State $t \sim \mathbb U[1, H]$			\Comment{Sample a switching index}
		\State $\tau_j = \text{Parse}(\mathcal D_j)$ \Comment{Parse the image, following $\pi_i$ till step $t$, and $\pi^*$ thereafter. }
		\State $G_t = G(\tau_{j}^t) $  \Comment{Evaluate returns at node $\tau_j^t$ via expert's cost to go in \cref{eq:returns}}.
		
		\State $\mathcal M \gets \mathcal M \cup \{(s_t,r_t,l_t,t,G_t)\}$ \Comment{Store the transition at index $t$ in memory}
		
		\EndFor																
		\State $\mathcal B \sim \mathbb U[\mathcal M], |\mathcal B| = B$ \Comment{Sample a minibatch from memory}
		\State $\theta \gets \theta + \alpha \nabla_{\theta}|_{\mathcal{B} }$ \Comment{Update $\theta$ via \cref{eq:grad7_acobj} approximated at $\mathcal B$}
		\State $\omega \gets \omega - \alpha \nabla_{\omega}|_{\mathcal{B}}$ \Comment{Update $\omega$ by gradient of objective in \cref{eq:critobj}  approximated at $\mathcal B$}
		\EndFor
	\end{algorithmic}
\end{algorithm}
However, as mentioned in section \ref{framingasrl}, maintaining a valid probability distribution over split locations in the stochastic policy case is challenging. 
We hence employ a \textit{deterministic} policy $\mu(s|\theta)$ for split locations --- making the stochsatic policy gradient update derived in \cite{Sun2017} unsuitable for learning $\mu(s|\theta)$. 
The AggreVaTe objective \cite{Ross2014} of minimizing the cost-to-go of the oracle in the deterministic policy setting may be expressed as: 
\begin{equation}
J_n(\theta) = \mathbb{E}_{\ t \sim U(1,...,H), s_t \sim d_{\mu_{1:n}}^t} \Big[ Q_t^*(s_t, \mu(s|\theta)) \Big].
\label{eq:newobj}
\end{equation}
Rather than sampling a split location $l_t$ from a stochastic policy, we retrieve the split location deterministically from the policy $l_t = \mu(s|\theta)$. 
As in \cite{Sun2017}, we may improve the policy by updating its parameters $\theta$ in the direction of improvement of $J_n(\theta)$, given by the gradient of equation \ref{eq:newobj}:
\begin{equation}
	\nabla_{\theta} J_n({\theta}) = \mathbb{E}_{\ t \sim U(1,...,H), s_t \sim d_{\mu_{1:n}}^t} \Big[ \nabla_{\theta} Q_t^* (s_t, \mu(s|\theta))\Big].
	\label{eq:gradobj}
\end{equation}
The Deterministic Policy Gradient \cite{Silver2014} allows us to evaluate this gradient, applying the chain rule:
\begin{equation}
	\nabla_{\theta} J_n({\theta}) = \mathbb{E}_{\ t \sim U(1,...,H), s_t \sim d_{\mu_{1:n}}^t} \Big[ \nabla_{a} \left. Q_t^* (s_t, a) \right\vert_{a = \mu(s|\theta)} \nabla_{\theta} \mu(s| \theta) \Big].
	\label{eq:graddpgt}
\end{equation}
The AggreVaTeD framework \cite{Sun2017} uses Monte Carlo samples $G_t$ of the oracle's cost to go, directly estimating  $Q_t^*(s_t,a)$ by sampling. 
While this provides an unbiased estimate of $Q_t^*$, we cannot compute the gradient $\nabla_a Q_t^*(s_t,a)$ using non-differentiable samples of the oracle's cost-to-go. 
Instead, we construct a \textit{differentiable} approximation of the oracle's cost to go, in the form of a critic network $Q(s,a | \omega)$, parametrized by $\omega$. 
While the notion of the critic network is similar to that present in DPG \cite{Silver2014} and DDPG \cite{Lillicrap2015}, note that our critic network approximates the cost-of-go of the \textit{oracle}, rather than the \textit{learner's} policy $\mu$, i.e., the critic is trained to optimize:
\begin{equation}
	\min_{\omega} \mathbb{E}_{s_t \sim d_{\mu_{1:n}}^t, a = \mu_{n}(s_t|\theta)} \Big[ \big( Q(s_t, a | \omega) - G_t  \big)^2 \Big].
	\label{eq:critobj}
\end{equation}
Using a critic network $Q_t(s_t,a|\omega)$ to approximate the oracle's cost-to-go
allows us to perform an update to the policy $\mu(s|\theta)$ by replacing $Q_t^*(s_t,a)$ in equations \ref{eq:gradobj} and \ref{eq:graddpgt} with critic network's estimate, $Q_t(s_t,a|\omega)$, leading to the following deterministic policy gradient update:
\begin{equation}
\nabla_{\theta} J_n({\theta}) = \mathbb{E}_{\ t \sim U(1,...,H), s_t \sim d_{\mu_{1:n}}^t} \Big[ \nabla_{a} \left. Q_t (s_t, a | \omega) \right\vert_{a = \mu(s|\theta)}  \nabla_{\theta} \mu(s| \theta) \Big].
\label{eq:grad_ddpgagg}
\end{equation}
Employing a differentiable critic network to estimate the oracle's cost-to-go thus introduces an \textit{deterministic actor critic variant of AggreVaTeD}, which we refer to as \DeterAggns. \DeterAgg hence serves as a deterministic policy gradient update that we use for training our deterministic split location policy $\mu(s|\theta)$ in the cost sensitive imitation learning setting.

We further note that applying the Deterministic Policy Gradient Theorem \cite{Silver2014} typically requires an approximation of the true gradient $\nabla_{\theta} J_n(\theta)$ from \cite{DBLP:journals/corr/abs-1205-4839}, due to the implicit dependence of $Q(s_t, a_t | \omega)$ on the parameters of the policy $\theta$. However, in \DeterAggns, the learned estimates of $Q(s_t, a_t| \omega)$ estimate the cost-to-go of the \textit{oracle} $\pi^*$, and not the learner's policy $\pi(a_t|s_t, \theta)$. The true cost-to-go of the oracle $Q^*(s_t,a_t)$ (and any estimate $Q(s_t,a_t | \omega)$ of this cost) are both \textit{independent} of the parameters $\theta$ of the learner's policy. 
\DeterAgg hence removes the dependence of $Q(s_t,a_t|\omega)$ on the learner's policy  $\theta$ by virtue of following the oracle after time $t$, hence the gradient update we present in equation \ref{eq:grad_ddpgagg} is no longer an approximation to the true gradient based on \cite{DBLP:journals/corr/abs-1205-4839}.

\DeterAgg allows us to compute a gradient update to deterministic split-location component of our neural parser, while we employ a standard stochastic actor-critic policy gradient update to for the rule policy $\pi(a|s,\theta)$.
This leads to a \textit{mixed} stochastic-deterministic policy gradient update in \cref{eq:grad7_acobj}:
\begin{multline}
\nabla_{\theta} J_n(\theta) = \mathbb{E}_{\ t \sim U(1,...,H), s_t \sim d_{\pi_{1:n}, \mu_{1:n}}^t, r_t \sim \pi_n(r| s_t, \theta)} \Big[ \nabla_{\theta} \log \pi(r_t | s_t, \theta) \cdot Q_t(s_t,r_t,\mu(s_t|\theta)|\omega)  \\ + \nabla_{l} \left. Q_t(s_t,r_t,l|\omega) \right\vert_{l = \mu(s_t|\theta)} \cdot \nabla_{\theta} \mu(s_t | \theta)\Big].
\label{eq:grad7_acobj}
\end{multline}
The full derivation of this mixed policy gradient update is provided in the supplementary material. 
The resultant \textit{mixed} stochastic-deterministic policy gradient update is based on \DeterAggns, and is applicable to any policy with both a deterministic and stochastic component. 
We utilize this mixed update to train our neural parser, as described in Algorithm \ref{alg}.

%% file: 3_newresults.tex
\section{Experimental Results and Analysis}
\vspace*{-1em}
\setlength{\textfloatsep}{0.25cm}
\begin{table}[t]
\centering
\caption{Parsing Accuracies of Proposed model and various baselines.}
\label{table}
\begin{tabular}{@{}lccc@{}}
\toprule
Model & Train Accuracy &  Test Accuracy \\ 
\midrule
 IGM Oracle with GT Access (Depth 7) & 98.50\% & --- \\ 
\midrule
Monte-Carlo Policy Gradient (RL) & 53.54\% & 51.23\% \\ 
DDPG (RL) & 51.94\% & 48.78\% \\
Behavior Cloning (IL) & 75.11\% & 75.10\% \\ 
DAgger (IL) & 84.01\% & 84.03\% \\
Stochastic  AggreVaTeD (IL+RL) & 81.85\% & 81.07\% \\ 
Actor-Critic AggreVaTeD (IL+RL) & 82.18\% & 81.31\% \\
Off-Policy Monte-Carlo  Policy Gradient (IL+RL) & 84.85\% & 83.65\% \\ 
Off-Policy Actor-Critic Policy Gradient (IL+RL) & 80.91\% & 80.94\% \\
\DeterAgg (IL+RL) \textbf{(Ours)} & \textbf{88.05\%} & \textbf{86.86\%} \\ \bottomrule
\end{tabular}
\end{table}
We evaluate our idea of learning to parse objects by imitation, 
and quantify how well our proposed \DeterAgg approach is able to parse \textit{novel} objects relative to baseline imitation learning (IL), reinforcement learning (RL) and hybrid IL+RL approaches, and the IGM oracle. 
This also provides us insight into the relative benefits of IL, RL, and IL+RL approaches in the context of parsing. The IL and IL+RL baselines are provided access to the IGM Oracle, with a maximum allowed parse tree depth of $7$ enforced for computational reasons. The considered baselines are as follows (with details in the supplementary).\\
\textbf{RL baselines:}
We consider two RL baselines where the learner must maximize its own cumulative reward without IGM oracle access: 1) An on-policy stochastic \textit{Monte-Carlo Policy Gradient (MCPG)} as in \cite{Williams1992}, 
2) A \textit{Deterministic Policy Gradient (DPG)} as in \cite{Lillicrap2015}. 
\\
\textbf{IL baselines:}
We consider two IL baselines, where the learner imitates the IGM oracle, with no reward function access. 1) \textit{Behavior Cloning (BC)}, where the agent imitates a  fixed set of demonstrated parses from the oracle, 
2) the interactive IL paradigm \textit{DAgger} \cite{Ross2011}. 
\\
\textbf{Hybrid IL+RL baselines:}
We consider four hybrid IL+RL baselines, where the learner has access to both the IGM oracle and the reward function during training. 
1) The original stochastic \textit{AggreVaTeD} policy gradient \cite{Sun2017}, 2) An  \textit{Actor-Critic variant of AggreVaTeD (AC-AggreVaTeD)},
3) A stochastic \textit{Off-Policy Monte-Carlo Policy Gradient (Off-MCPG)}, where the oracle acts as a behavioral policy, and it's actor-critic variant, 
4) An \textit{Off-Policy Actor-Critic Policy Gradient (Off-ACPG)}.
Note that all stochastic baselines are evaluated selecting the most likely action at test time. 

\begin{figure*}[t]
	\centering
	\includegraphics[width=\linewidth]{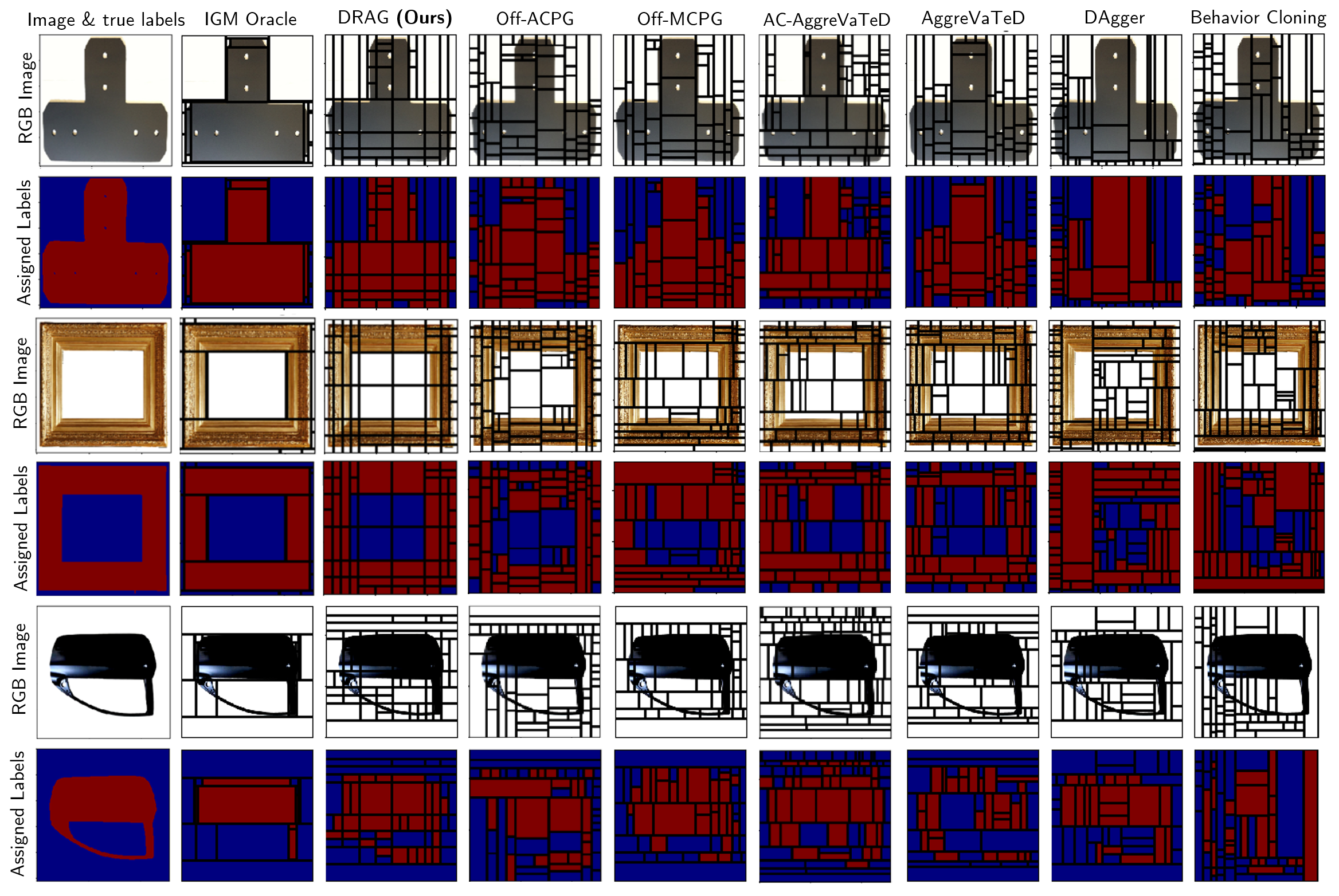}
	\vspace*{-2em}
	\caption{ \small
	Depiction of constructed sample parses for a metal plate (rows 1 \& 2), a window frame (rows 3 \& 4), and a car door (rows 5 \& 6), from the oracle (column 2), the proposed \DeterAgg (column 3), and the various baselines (columns 4-9). The first column shows the original image (odd rows) with ground truth labels (even rows, red object pixels are to be painted, blue are not to be painted). Each column shows the segmented object image along with the predicted label assignment.}
	\label{fig:results}
\end{figure*}

\textbf{Experimental Setup:} To evaluate our proposed \DeterAgg approach against the above baselines, we collect a set of $362$ RGB object images of size $256 \times 256$ pixels, each annotated with a per-pixel binary label of $1$ (to be painted), or $−1$ (not to be painted), serving as ground truth object labels.
We evaluate our models with $3$-fold cross validation, training on $3$ sets of $300$ randomly sampled images, measuring performance on $3$ corresponding sets of $62$ test images.
We measure performance as the pixel accuracy between the predicted assignment of labels of the object image against these ground truth object labels, presented for each of the above baselines and our approach in Table \ref{table}. 
An ideal parse assigns contiguous paint labels to all portions of the object to be painted, ensuring that parts of the object do not go unpainted.
We present the parses created for $3$ sample images in \cref{fig:results}.

\textbf{Analysis:} We make the following observations based on the results in Table \ref{table} and Figure \ref{fig:results}: \\
    \textit{1) Reinforcement learning applied to our task is unsuccessful.}
    The recursive nature of the parsing process requires good rules and split locations to be selected \textit{consistently} for a good parse, which is unlikely via random exploration. 
    The MCPG and DPG baselines rely on random exploration of rules and splits, thus only achieving random performance and failing to parse object images at all.
    
\textit{2) Imitation learning applied to our parsing task is successful.} 
    The structural similarity between IGM decision-trees and parsing allows the IGM oracle to \textit{guide sampling} towards good rules and split locations rather than the naive exploration of RL.
    As seen in \cref{table}, even our most naive IL baseline, Behavior Cloning, significantly outperforms the RL baselines, achieving $75.10\%$ test accuracy. DAgger reconciles with the state distribution mismatch \cite{Ross2011}, boosting the parser to $84.03\%$ test accuracy. 
    The IL paradigm is thus able to learn reasonable parses of object images, overcoming the issues of learning a parser present in the RL setting.
    
\textit{3) \DeterAgg outperforms state-of-the-art IL and RL baselines on the task of parsing novel objects.}
    Quantitatively, our proposed \DeterAgg approach achieves notably higher train and test accuracies (88.05 \% and 86.86\% respectively) than other baseline approaches. 
We note the following:
\begin{enumerate}[leftmargin=0.2in,noitemsep,label=(\alph*),topsep=-2pt]
    \item While some improvement of performance may be attributed to the use of a lower variance return estimate offered by a critic network, comparing the performance achieved by the actor-critic baselines against their actor-only variants shows there is little benefit to using a critic network with a stochastic policy representation.
    The deterministic policy representation used in \DeterAgg allows for better learning of split locations, contributing towards the notable improvement in performance observed. 
    
    \item The deterministic policy representation of split locations used in \DeterAgg enables it to construct more regular parses of the object images by selecting splits that align with object boundaries,  as compared to baseline approaches. This is particularly suitable for axis-aligned images, or images with small aberrations (such as rows 3 \& 4 in \cref{fig:results}).
    While the stochastic hybrid IL+RL baselines are able to capture coarse object structure, they fail to capture splits aligned with prominent image gradients.
    In contrast, \DeterAgg is able to assign labels that correlate strongly with that of the ground truth (and the IGM oracle). 
    
    \item \DeterAgg better optimizes for the underlying cost compared to other hybrid IL+RL baselines, generalizing past demonstrated expert actions. This is exemplified in the case of irregular images such as the car door in rows 5 \& 6 of \cref{fig:results}. Here, the IGM oracle misses out on the correctly labeling the rim of the door due to a restricted parse tree depth. Despite this lack of supervisory actions to imitate, \DeterAgg learns to correctly label this portion of the door, and is capable of performing cost-sensitive imitation in a superior manner compared to other baseline approaches.
\end{enumerate}


%% file: 4_conclusion.tex
\section{Conclusion}
\vspace*{-1em}
In this paper, we address the problem of learning to parse objects into hierarchical decompositions via imitation learning.  
By treating an Information Gain Maximizing algorithm as an expert parsing oracle, our neural parser learns to parse objects by imitating this IGM oracle, observing only raw object images as input. 
We further introduce a novel deterministic policy gradient update, \DeterAgg, suitable for generic imitation learning tasks with a deterministic policy representation. 
The proposed \DeterAgg may be seen as both as an deterministic actor-critic variant of AggreVaTeD \cite{Sun2017} and a variant of DDPG \cite{Lillicrap2015} suitable for imitation learning. 
Training our neural parser to parse objects using \DeterAgg outperforms existing RL, IL and IL+RL baselines, leading to more accurate and coherent parses. Our experimental results demonstrate the capability of our approach to successfully parse objects, and potentially address more generic spatially decomposable tasks. 



%% file: supplementary_body.tex
\section*{Supplementary Material}

\subsection*{Mixed Policy Gradient:}
In section 3.4 of the main paper, we derived a deterministic policy gradient update to deterministic policy trained in the AggreVaTeD \cite{Sun2017} setting. 


The specific policy representation that we employ in our neural parser has a stochastic component predicting which rules to apply, $\pi(r|s_t, \theta)$, as well as a deterministic component to predict split locations, $\mu(s_t | \theta)$. Given such a \textit{mixed} stochastic-deterministic policy representation, we derive a corresponding \textit{mixed} stochastic-deterministic policy gradient update that we employ to train our neural parser.

Consider that at any time step $t$, rules $r_t$ are sampled from $\pi(r|s_t, \theta)$, and split locations $l_t$ are given by $\mu(s_t|\theta)$. Following the AggreVaTeD \cite{Sun2017} training paradigm, we seek to train the components of our policy $\pi(r|s_t, \theta)$ and $\mu(s_t|\theta)$, to maximize the cost to go of the expert, $Q^*(s_t, r_t, l_t)$. Formally, we seek to maximize: 
\begin{equation}
J_n(\theta) = \mathbb{E}_{\ t \sim U(1,...,H), s_t \sim d_{\pi_{1:n}, \mu_{1:n}}^t, r_t \sim \pi_n(r|s_t, \theta)} \Big[ Q_t^*(s_t,r_t,\mu(s_t|\theta)) \Big]
\label{eq:mcobj}
\end{equation}
As described in section 3.4 of the main paper, the actor critic variant of AggreVaTeD uses a learnt \textit{estimate} $Q(s_t, r_t, l_t | \omega)$ of the cost-to-go of the expert $Q^*(s_t,r_t,l_t)$. The objective equation \ref{eq:mcobj} thus becomes:
\begin{equation}
J_n(\theta) = \mathbb{E}_{\ t \sim U(1,...,H), s_t \sim d_{\pi_{1:n}, \mu_{1:n}}^t, r_t \sim \pi_n(r|s_t, \theta)} \Big[ Q_t(s_t,r_t,\mu(s_t|\theta)|\omega) \Big]
\label{eq:acobj}
\end{equation}
The expectation of $r_t \sim \pi(r| s_t , \theta)$ may be represented as follows:
\begin{equation}
J_n(\theta) = \mathbb{E}_{\ t \sim U(1,...,H), s_t \sim d_{\pi_{1:n}, \mu_{1:n}}^t} \Big[ \sum_{r \in \mathcal{R}} \  \pi(r_t | s_t, \theta) \ Q_t(s_t,r_t,\mu(s_t|\theta)|\omega) \ \Big]
\label{eq:exp_acobj}
\end{equation}

To compute an update to the policy, we may compute the gradient of this objective $J_n(\theta)$ with respect to the parameters of the policy $\theta$: 
\begin{equation}
\nabla_{\theta} J_n(\theta) = \nabla_{\theta} \ \mathbb{E}_{\ t \sim U(1,...,H), s_t \sim d_{\pi_{1:n}, \mu_{1:n}}^t} \Big[ \sum_{r \in \mathcal{R}} \  \pi(r_t | s_t, \theta) \ Q_t(s_t,r_t,\mu(s_t|\theta)|\omega) \ \Big]
\label{eq:grad1_acobj}
\end{equation}
Considering the linearity of expectations, and taking the gradient $\nabla_{\theta}$ inside the sum $\sum_{r \in \mathcal{R}}$, this gives us: 
\begin{equation}
\nabla_{\theta} J_n(\theta) = \mathbb{E}_{\ t \sim U(1,...,H), s_t \sim d_{\pi_{1:n}, \mu_{1:n}}^t} \Big[  \sum_{r \in \mathcal{R}} \nabla_{\theta} \Big\{  \pi(r_t | s_t, \theta) \ Q_t(s_t,r_t,\mu(s_t|\theta)|\omega) \Big\} \Big]
\label{eq:grad2_acobj}
\end{equation}
Note that Proposition (1) from \citet{905687} shows that the gradient of a cummulative reward objective, $\nabla_{\theta} J_n(\theta)$, is independent of the gradient of the state distribution $d^t_{\pi_{1:n},\mu_{1:n}}$. 
Applying the product rule, we have: 
\begin{multline}
\nabla_{\theta} J_n(\theta) = \mathbb{E}_{\ t \sim U(1,...,H), s_t \sim d_{\pi_{1:n}, \mu_{1:n}}^t} \Big[  \sum_{r \in \mathcal{R}} \Big\{  \nabla_{\theta} \pi(r_t | s_t, \theta)  \ . \ Q_t(s_t,r_t,\mu(s_t|\theta)|\omega)  \\ + \pi(r_t | s_t, \theta) \ . \nabla_{\theta} Q_t(s_t,r_t,\mu(s_t|\theta)|\omega) \Big\} \Big]
\label{eq:grad3_acobj}
\end{multline}
The first term in the expectation, $\nabla_{\theta} \pi(r_t | s_t, \theta)  \ . \ Q_t(s_t,r_t,\mu(s_t|\theta)|\omega)$ may be simplified by applying the importance sampling trick:
\begin{multline}
\nabla_{\theta} J_n(\theta) = \mathbb{E}_{\ t \sim U(1,...,H), s_t \sim d_{\pi_{1:n}, \mu_{1:n}}^t} \Big[  \sum_{r \in \mathcal{R}} \Big\{  \pi(r_t | s_t, \theta) \ . \frac{\nabla_{\theta} \pi(r_t | s_t, \theta)}{\pi(r_t | s_t, \theta)}  \ . \ Q_t(s_t,r_t,\mu(s_t|\theta)|\omega)  \\ + \pi(r_t | s_t, \theta) \ . \nabla_{\theta} Q_t(s_t,r_t,\mu(s_t|\theta)|\omega) \Big\} \Big]
\label{eq:grad4_acobj}
\end{multline}
This becomes the gradient of the \textit{log-probability} of the policy $\pi(r|s_t, \theta)$:
\begin{multline}
\nabla_{\theta} J_n(\theta) = \mathbb{E}_{\ t \sim U(1,...,H), s_t \sim d_{\pi_{1:n}, \mu_{1:n}}^t} \Big[  \sum_{r \in \mathcal{R}} \Big\{  \pi(r_t | s_t, \theta) \ . \nabla_{\theta} \log \pi(r_t | s_t, \theta) \ . \ Q_t(s_t,r_t,\mu(s_t|\theta)|\omega)  \\ + \pi(r_t | s_t, \theta) \ . \nabla_{\theta} Q_t(s_t,r_t,\mu(s_t|\theta)|\omega) \Big\} \Big]
\label{eq:grad5_acobj}
\end{multline}
The second term in the expectation, $\pi(r_t | s_t, \theta) \ . \nabla_{\theta} Q_t(s_t,r_t,\mu(s_t|\theta)|\omega)$, may be computed using the Deterministic Policy Gradient Theorem \cite{Silver2014} (i.e. essentially applying the chain rule): 
\begin{multline}
\nabla_{\theta} J_n(\theta) = \mathbb{E}_{\ t \sim U(1,...,H), s_t \sim d_{\pi_{1:n}, \mu_{1:n}}^t} \Big[  \sum_{r \in \mathcal{R}} \Big\{  \pi(r_t | s_t, \theta) \ . \nabla_{\theta} \log \pi(r_t | s_t, \theta) \ . \ Q_t(s_t,r_t,\mu(s_t|\theta)|\omega)  \\ + \pi(r_t | s_t, \theta) \ . \nabla_{l} \left. Q_t(s_t,r_t,\mu(s_t|\theta)|\omega) \right\vert_{l = \mu(s_t|\theta)} \ . \nabla_{\theta} \mu(s_t | \theta) \Big\} \Big]
\label{eq:grad6_acobj}
\end{multline}
We may now convert the expression $ \sum_{r \in \mathcal{R}} \pi(r|s_t, \theta)$ back to an expectation of $r_t \sim \pi(r| s_t , \theta)$, leading to the following policy gradient update to our neural parser:
\begin{multline}
\nabla_{\theta} J_n(\theta) = \mathbb{E}_{\ t \sim U(1,...,H), s_t \sim d_{\pi_{1:n}, \mu_{1:n}}^t, r_t \sim \pi_n(r| s_t, \theta)} \Big[ \nabla_{\theta} \log \pi(r_t | s_t, \theta) \ . \ Q_t(s_t,r_t,\mu(s_t|\theta)|\omega)  \\ + \nabla_{l} \left. Q_t(s_t,r_t,\mu(s_t|\theta)|\omega) \right\vert_{l = \mu(s_t|\theta)} \ . \nabla_{\theta} \mu(s_t | \theta)\Big]
\label{eq:grad7_acobj}
\end{multline}
Note that applying the Deterministic Policy Gradient Theorem typically requires an approximation from \cite{DBLP:journals/corr/abs-1205-4839}, due to the implicit dependence of $Q(s_t, a_t | \omega)$ on the parameters of the policy $\theta$. However, in the deterministic variant of AggreVaTeD, the learnt estimates of $Q(s_t, a_t| \omega)$ estimate the cost-to-go of the \textit{expert policy} $\pi^*$, and not the learner's policy $\pi(a_t|s_t, \theta)$. 

The true cost-to-go of the expert $Q^*(s_t,a_t)$ (and any estimate $Q(s_t,a_t | \omega)$ of this cost) are both \textit{independent} of the parameters of the policy $\theta$. The deterministic actor-critic variant of AggreVaTeD removes the dependence of $Q(s_t,a_t|\omega)$ on $\theta$, thus removes the necessity for the approximation from \citet{DBLP:journals/corr/abs-1205-4839}.


\subsection*{Use of a Memory Replay:}
Uniformly sampling from a replay memory (as in Algorithm 1) corresponds to sampling states from the distribution of states encountered during training, i.e. a sample approximation of the aggregated state distribution $d^t_{\pi_{1:n},\mu_{1:n}}$, justifying the use of a replay memory in the aggregated state distribution case. Indeed, using states visited from training iterations $1$ to $n$ is closer to the original AggreVaTe objective \cite{Ross2014} than AggreVaTeD \cite{Sun2017}, which uses only states from iteration $n$.

\subsection*{Details of Baseline Algorithms:}
We describe the exact policy representation used in each of the baseline algorithms mentioned in the main paper below. For ease of comparison, we also provide a table of the training setting used in the various baseline approaches and our model in \cref{table2}. Each model uses a convolutional neural network with $7$ convolutional layers and $2$ dense layers as a base model. The baseline approaches then represent their respective policies as follows:

\textbf{Pure RL baselines:} 
In the pure RL setting, the learner is provided with evaluations of the quality of the parses it constructs via the reward function, and does not have access to the IGM oracle agent in any form. We consider two RL baselines, where the objective is to simply maximize the cumulative reward achieved by the learner: 
\begin{itemize}[leftmargin=0.2in,itemsep=0.2pt,topsep=-2pt]
    \item \textit{Monte-Carlo Policy Gradient (MCPG):} We consider an on-policy stochastic Monte-Carlo Policy Gradient approach, similar to REINFORCE. The policy maintains a categorical distribution over valid rules, predicted as a softmax of deep network features over the valid rules applicable at the current image segment, and a logit-normal distribution over valid split locations within the boundaries of the current image segment.
    \item \textit{Deterministic Policy Gradient (DPG):} We then consider a DDPG \cite{Lillicrap2015} style approach, where the policy deterministically predicts split locations as scaled logistic function of the object image features. The rules here are still predicted stochastically as a softmax output of the deep policy network. 
\end{itemize}

\textbf{Pure IL baselines:}
In contrast with the RL setting, the learner in the pure IL setting has access to the actions taken by the IGM oracle, and not the reward function. Here, the learner simply tries to copy the actions executed by the expert; this corresponds to maximizing the likelihood of the rules selected by the expert under the learner's policy, and \textit{regressing} to the split locations selected by the expert. We consider two IL baselines:
\begin{itemize}[leftmargin=0.2in,itemsep=0.2pt,topsep=-2pt]
 \item \textit{Behavior Cloning:} Here, the agent minimizes the categorical cross entropy between the rules selected by the IGM oracle, and the softmax probability distribution over rules predicted by the policy network. This is equivalent to maximizing the log-likelihood of the rules selected by the IGM oracle. The split locations are predicted as a scaled logistic function of the deep network features. A L2 norm loss between the predicted and IGM oracle split is used to train the agent's split location policy.
 \item \textit{DAgger:} Following the interactive learning paradigm DAgger \cite{Ross2011}, objects are parsed according to a \textit{mixture} of the expert and the learner's current policy. The policy representation is identical to that used in the Behavior Cloning case. 
\end{itemize}

\textbf{Hybrid IL+RL baselines:}
Of particular interest to us is the hybrid IL+RL case, where the learner has access to both the actions executed by the expert, as well as samples of the reward function for the parses it constructs. 
\begin{itemize}[leftmargin=0.2in,itemsep=0.2pt,topsep=-2pt]
    \item \textit{AggreVaTeD:} We consider the original stochastic policy gradient training paradigm of AggreVaTeD \cite{Sun2017}. As in the Off-MCPG case, the rules are predicted via a categorical distribution from the deep policy network features, and the split locations are predicted as a logit-normal distribution over valid splits.
    \item \textit{Actor-Critic AggreVaTeD:} We consider an Actor-Critic variant of AggreVaTeD \cite{Sun2017}, where the Monte-Carlo estimate of the oracle's return is replaced by the critic's estimate of this return. The critic is trained using \cref{eq:critobj} as in \DeterAgg. \item \textit{Off-Policy Monte-Carlo Policy Gradient (Off-MCPG):}  
    Treating the IGM oracle as a behavioral policy, we maximize the learner's returns via an off-policy Monte-Carlo policy gradient.
    As in the case of the vanilla MCPG, the policy representation is a categorical distribution over the valid rules, and the splits with a logit-normal distribution over valid splits.
    \item \textit{Off-Policy Actor-Critic Policy Gradient (Off-ACPG):}
    We finally consider an actor-critic variant of Off-MCPG, using a critic network to estimate the oracle's return. As in the case of AC-AggreVaTeD, we use \cref{eq:critobj} to train the critic. 
\end{itemize}
\begin{table}[t]
\centering
\caption{Training Setting Ablation of Proposed model and various baselines.}
\label{table2}
\begin{tabular}{@{}lccc@{}}
\toprule
Model & Training Setting & Policy Representation &  Actor / Actor-Critic \\ 
\midrule
MCPG & RL & Stochastic & Actor \\ 
DDPG & RL & Deterministic & Actor-Critic \\
\midrule
Behavior Cloning & IL & Deterministic & Actor \\ 
DAgger & IL & Deterministic & Actor \\
\midrule
AggreVaTeD & IL+RL & Stochastic & Actor \\ 
AC AggreVaTeD & IL+RL & Stochastic & Actor-Critic \\
Off-MCPG & IL+RL & Stochastic & Actor \\ 
Off-ACPG & IL+RL &  Stochastic & Actor-Critic \\
\DeterAgg \textbf{(Ours)} & IL+RL & Deterministic & Actor-Critic \\ \bottomrule
\end{tabular}
\end{table}
\subsection*{Training Details and Choice of Hyperparameters:}
We note details regarding our training setup, as well as values of hyperparameters used during training the various baseline approaches and our model:
\begin{itemize}[topsep=-0.1in,leftmargin=0.2in]
    \item \textit{Convergence:} We ensure each approach is trained till convergence by evaluating the model on the test set after the validation accuracy saturated (is no longer improving significantly).
    \item \textit{3-fold Cross-Validation:} Our image based problem allows us to maintain distinct training and testing image sets, we follow 3-fold cross-validation, maintaining 3 different train-test sets, and reporting average train and test accuracies across these sets. 
    \item \textit{Learning Rate:} For all models, we use the Adam Optimizer available in TensorFlow \cite{tensorflow2015-whitepaper}. All our models are trained with an learning rate of $10^{-4}$.  
    \item \textit{Mixing Coefficient $\beta$:} For the DAgger baseline, as well as the hybrid IL+RL baselines and \DeterAggns, we utilize an initial mixing coefficent of $1$ (i.e. start off using the expert policy alone), annealed to a final value of $0.5$ (a $50\%$ chance to use either the learner's policy or the expert policy), over $100$ training epochs. 
    \item \textit{Gradient Clipping:} For all models, we apply a gradient clipping to a maximum value of $10$. 
\end{itemize}